\documentclass[draft]{agujournal2019}
\usepackage{url} 
\usepackage{lineno}
\usepackage[inline]{trackchanges} 
\usepackage{soul}
\usepackage{amsmath}
\usepackage{multirow}
\usepackage{array}
\draftfalse
\usepackage{booktabs}

\journalname{XXX}

\begin{document}

%
%


\title{Baseflow identification via explainable AI with Kolmogorov-Arnold networks}

%
%




\authors{Chuyang Liu\affil{1}, Tirthankar Roy\affil{2}, Daniel M. Tartakovsky\affil{3}\thanks{},  Dipankar Dwivedi\affil{1}}

\affiliation{1}{Lawrence Berkeley National Laboratory, Geochemistry Department, Berkeley, California, 94720, USA}
\affiliation{2}{University of Nebraska-Lincoln, Department of Civil and Environmental Engineering, Lincoln, Nebraska, 68588, USA}
\affiliation{3}{Department of Energy Science and Engineering, Stanford University, Stanford, California, 94035, USA}









\begin{keypoints}
\item Kolmogorov-Arnold networks (KANs) enhance interpretability of machine-learned hydrological models.

\item KAN-derived symbolic formulations outperform state-of-the-art semi-empirical aridity indices.
\item KAN-identified functional form yields an analytical index with fewer fitting parameters and improved performance.

\end{keypoints}

%
%

%
%


\begin{abstract}
Hydrological models often involve constitutive laws that may not be optimal in every application. We propose to replace such laws with the Kolmogorov-Arnold networks (KANs), a class of neural networks designed to identify symbolic expressions. We demonstrate KAN's potential on the problem of baseflow identification, a notoriously challenging task plagued by significant uncertainty. KAN-derived functional dependencies of the baseflow components on the aridity index outperform their original counterparts. On a test set, they increase the Nash-Sutcliffe Efficiency (NSE) by 67\%, decrease the root mean squared error by 30\%, and increase the Kling-Gupta efficiency by 24\%. This superior performance is achieved while reducing the number of fitting parameters from three to two. Next, we use data from 378 catchments across the continental United States to refine the water-balance equation at the mean-annual scale. The KAN-derived equations based on the refined water balance outperform both the current aridity index model, with up to a 105\% increase in NSE, and the KAN-derived equations based on the original water balance. While the performance of our model and tree-based machine learning methods is similar, KANs offer the advantage of simplicity and transparency and require no specific software or computational tools. This case study focuses on the aridity index formulation, but the approach is flexible and transferable to other hydrological processes.

\end{abstract}

\section*{Plain Language Summary}

Equations used in hydrologic model are often suboptimal, resulting in reduced prediction accuracy and efficiency. We implemented Kolmogorov-Arnold networks (KAN), a machine learning algorithm for deriving symbolic formulations, to estimate groundwater recharge and showed that it outperforms an existing state-of-the-art semi-empirical formulation. In hydrology, Nash-Sutcliffe Efficiency (NSE), Root Mean Squared Error (RMSE), and Kling-Gupta Efficiency (KGE) are commonly used to evaluate model performance. Higher NSE and KGE values indicate better performance, while lower RMSE values are preferable. Our results show that NSE increased by 71\%, RMSE decreased by 32\%, and KGE improved by 25\%. In addition, KAN is capable of identifying an optimal functional form, and KAN can be used to drive a new analytical formula using the prior knowledge. The KAN-inspired equation outperformed the original formulation and reduced the fitting parameters. Furthermore, we refined the water-balance equation at the mean-annual scale and showed that, based on the new water-balance equation, KAN can derive new formulations that are superior to the original aridity index formulations (up to 105\% increase in NSE) and KAN-derived equations based on the original water balance. These findings highlight the significant potential of KAN to advance the scientific understanding of a wide range of hydrologic processes.

%
%

%


%
%
%
%

\section{Introduction}

Hydrological models are used to quantify the water cycle and its interactions with human activities and ecosystems. They provide crucial insights into water resource dynamics and support decision-making in various applications. Data-driven construction of such models, or empirical constitutive laws therein, typically involves either parametric (non-symbolic) or symbolic regression. Non-symbolic regression fits data to a predefined model, while symbolic regression discovers the best-fitting equations from the data. Examples of non-symbolic regression include the curve number method  \cite{ponce1996runoff}, which uses an empirical formula to relate runoff to land use, soil type, and hydrological conditions; the Penman-Monteith equation to estimate evapotranspiration from multiple meteorological and physiological factors \cite{allen1998crop}; and various functional forms of the moisture retention curve \cite{assouline-2001-unsaturated} that relates soil saturation to pore pressure. Such pre-defined constitutive relations introduce bias due to their failure to capture complex and dynamic relationships between relevant hydrologic variables. They also have a limited ability to generalize across diverse hydrological conditions and ecosystems.

Symbolic regression aims to discover new functional relationships contained in the data. A prime example from this class is a deep neural network (NN), a universal approximator of any continuous function. It yields black-box models, aka equation-free constitutive relations, which are hard to interpret. NNs-based symbolic regression ameliorates this limitation by yielding a closed-form formula for various constitutive laws \cite{martius2016extrapolation, sahoo2018learning, zhou2022bayesian, Kubal_k_2023}. Such symbolic NNs can handle binary (e.g., addition, subtraction) and unary (e.g., exponential, sine, logarithm) operations at different units. The significance of each mathematical operation is represented by the magnitude of its weight, whose magnitude is adjusted, together with the biases of neurons, during the NN training. The combination of these optimized weights and operations gives the final formula. These approaches face challenges, such as the algorithm's failure to find an accurate model, making it difficult to interpret intermediate processes and to identify errors.

The Kolmogorov-Arnold networks (KANs) address this challenge by leveraging the Kolmogorov-Arnold representation theorem. The latter states that any continuous function can be represented as a combination of continuous functions of a single variable \cite {liu2024kan}. Like traditional NNs, KANs use gradient-based optimization for training and can handle various mathematical operations. While the traditional NN architecture utilizes a fixed activation function, KAN's architecture is designed to learn activation functions that fit the data best. The key advantage of KAN is its ability to visualize learned activation functions and to provide interpretable symbolic representations of these functions with prior knowledge.

We demonstrate the value of KANs by identifying optimal functional dependencies between the runoff components and the aridity index. The latter is defined as the ratio of mean annual potential evapotranspiration to mean annual precipitation. It represents regional dryness and its impact on water balance. This well-established indicator has significant implications for hydrological modeling. For example, semi-empirical models were used to estimate long-term (mean-annual) direct runoff and baseflow from the aridity index across 378 catchments in the contiguous United States \cite{aridity}.

By providing more precise functional dependencies on the aridity index, we aim to improve the accuracy and interpretability of hydrological models. To achieve this, our research addresses two key questions. First, can a KAN identify novel functional dependencies on the aridity index? Second, how do the KAN-derived symbolic relations involving the aridity index compare with their current counterparts in terms of accuracy, simplicity, and predictive power? By exploring these questions, we demonstrate that KANs can yield superior models in terms of accuracy and simplicity, highlighting their potential to improve hydrological modeling across various applications.

\section{Materials and Methods}

\subsection{Datasets}

The Catchment Attributes and Meteorology for Large-sample Studies (CAMELS) dataset \cite{addor2017camels} includes daily streamflow, meteorology, and other attributes for 671 catchments across the contiguous United States. Using this dataset, \citeA{aridity} calculated the mean-annual precipitation, $P$; the mean-annual potential evapotranspiration, PET; the mean-annual direct runoff, $Q_D$; the mean-annual baseflow, $Q_B$; and the aridity index $\phi = \text{PET} / P$. Briefly, after excluding catchments that did not meet the criteria for data completeness, catchment size, snow fraction, and positive annual evaporation, 378 out of 671 CAMELS catchments were selected for analysis. Daily streamflow data were partitioned into direct runoff and baseflow using the low-pass filter method \cite{lyne1979hydrology}. Daily potential evapotranspiration was calculated using the Reference-crop Penman-Monteith formulation, based on incoming solar radiation and wind speed at two meters from the gridMET dataset \cite{abatzoglou2013development}, along with other variables from the CAMELS dataset. Finally, daily values were aggregated to mean-annual variables for each catchment. The processed dataset is available at the repository \cite{Meira2020}. We use this dataset in our analysis.

\subsection{Aridity Index Formulations}
Aridity index formulations of \citeA{aridity} are derived using the \citeA{budyko1974climate} and the \citeA{lvovich1979world} frameworks. In the former, $P$ is decomposed into mean annual streamflow, $Q$, and evaporation, $E$, such that 
\begin{equation}\label{eq:1}
\frac{E}{P} = 1 - \frac{Q}{P} 
\equiv f_E(\phi),
\end{equation} 
where the function $f_E$ describes a relationship between $E / P$ and $\phi$ (aridity index).

In the \citeA{lvovich1979world} framework, annual precipitation $p$ is partitioned into annual direct runoff $q_D$, annual baseflow $q_B$, and annual evapotranspiration $e$. 
At the mean-annual time scale, this partition is written as 
\begin{equation}\label{eq:2}
\frac{Q}{P} = \frac{Q_D}{P} + \frac{Q_B}{P}.
\end{equation}
From Eq.~\eqref{eq:1}, $Q / P$ is a function of $\phi$.
Assuming that both terms on the right-hand-side of Eq.~\eqref{eq:2} are also functions of $\phi$, 
\begin{subequations}\label{eq:Pnorm_all}
\begin{equation} 
\frac{Q}{P} = f_D(\phi) + f_B(\phi),
\end{equation} 
where 
\begin{equation} \label{eq:Pnorm_fdl}
f_D(\phi) = \frac{Q_D}{P} \quad\text{and}\quad
f_B(\phi) = \frac{Q_B}{P}.
\end{equation}
\end{subequations}

Based on the observed patterns of $f_D(\phi)$ and $f_B(\phi)$ in the aforementioned datasets, \citeA{aridity} assigned to these functions an exponential form,
\begin{equation} \label{eq:exp_decay}
f_i(\phi) = \exp(-\phi^{a_i} + b_i)^{c_i}, \qquad i = D, B.
\end{equation} 
where $a_i$, $b_i$, and $c_i$ are the fitting parameters in the $i$th function. 
After randomly splitting the datasets into halves to fit parameters and evaluate model performance 100 times, \citeA{aridity} obtained
\begin{equation} \label{eq:fb_org}
\frac{Q_B}{P} \equiv f_B(\phi) = \exp(-\phi^{1.71} - 0.873)^{1.05}
\end{equation}
and
\begin{equation} \label{eq:fd_org}
\frac{Q_D}{P} \equiv f_D(\phi) = \exp(-\phi^{0.77} - 0.864)^{1.06}.
\end{equation}
These relations capture the significant variability in $Q_D / P$ and $Q_D / P$ across 378 catchments in the contiguous US \cite{aridity}. 

\subsection{Development of KAN Models}

We train KAN models on the same dataset, which was randomly split into a training set and a test set with an 8 to 2 ratio. To avoid overfitting, we used a 10-fold cross-validated grid-search method \cite{pedregosa2011scikit}. In this method, the training dataset was split into a pre-training set (any nine folds out of the ten folds) and a validation set (the remaining one fold out of the ten folds). The pre-training set was used to develop a KAN with a given set of hyperparameters, and the validation set was used to evaluate the model performance by the coefficient of determination, R$^2$. The Broyden Fletcher Goldfarb Shanno (BFGS) algorithm \cite{head1985broyden} was used to find model parameters that minimize the loss, defined as the Root Mean Square Error (RMSE) \cite{chai2014root}. The grid search in this method iterated through all sets of hyperparameters and each set of hyperparameters has an average R$^2$. The optimal set of hyperparameters was chosen based on the highest average $\mathrm{R^2}$. The model was trained with the optimal hyperparameters and the training set was selected as the final KAN model.

In the 10-fold cross-validation with a single set of hyperparameters, ten KAN models were developed. Each KAN model was developed through five steps. In step 1, a fully connected KAN is initially trained to be sparse by regularization. To retain significant connections, pruning is used in step 2 to remove insignificant ones. Based on learned activation functions in the remaining connections, symbolic functions are set in step 3 through the optimal curve-fitting with the highest $\mathrm{R^2}$. In step 4, affine parameters are further refined by additional training. In the final step, the combination of trained parameters and symbolic functions is expressed as a symbolic formula. After training each KAN model, the ten $\mathrm{R^2}$ values were averaged to obtain the average $\mathrm{R^2}$ for a single set of hyperparameters. In this study, hyperparameters, including the number of layers, the number of neurons in each layer, the seed number to initialize the model, and the number of grid intervals for each spline, are evaluated.

\subsubsection{Kolmogorov-Arnold Networks}

The Kolmogorov-Arnold representation theorem states that a continuous multivariate function on a bounded domain can be expressed as a summation of finite univariate functions \cite{Kolmogorov, Griebel}. Based on this theorem, \citeA{liu2024kan} proposed a NN architecture called KAN as an alternative to Multi-Layer Perceptrons (MLPs) with arbitrary widths and depths. Unlike MLPs, which have fixed activation functions, KAN utilizes learnable univariate functions parameterized by B-splines. Each activation function in a KAN can be set symbolically, forming the final symbolic formula.

Given an input vector $\mathbf{x}$, the output of a KAN with $L$ layers is the composition of each layer:
\begin{equation}
\text{KAN}(\mathbf{x}) = (\Psi_{L-1} \circ \Psi_{L-2} \circ \cdots \circ \Psi_1 \circ \Psi_0)\mathbf{x},
\end{equation}
where $\Psi_{l}$ is the activation function matrix for the $l$th KAN layer.
In a KAN, neuron $i$ at layer $l$ is denoted by  $(l,i)$, neuron $j$ at layer $l+1$ is denoted by $(l+1,j)$, and the activation function connecting $(l,i)$ and $(l+1,j)$ is denoted by $\psi_{l,j,i}$. The input to neuron $(l+1,j)$ is the summation of all post-activation values from connected neurons at layer $l$:
\begin{equation}
x_{l+1,j} = \sum_{i=1}^{n_l} \psi_{l,j,i}(x_{l,i}), \quad j = 1, \cdots, n_{l+1}
\end{equation}
The output of a KAN is written as
\begin{align}
\begin{split}
y = & \; \text{KAN}(\mathbf{x}) \\
= & \sum_{i_L=1}^{n_L-1} \psi_{L-1,i_L,i_{L-1}} \left( \sum_{i_{L-2}=1}^{n_{L-2}} \cdots \left( \sum_{i_2=1}^{n_2} \psi_{2,i_3,i_2} \left( \sum_{i_1=1}^{n_1} \psi_{1,i_2,i_1} \left( \sum_{i_0=1}^{n_0} \psi_{0,i,i_0}(x_{i_0}) \right) \right) \right) \cdots \right).
\end{split}
\end{align}
The activation function $\psi(x)$ is defined as
\begin{equation}
\psi(x) = w_b \frac{x}{1+\text{e}^{-x}} + w_c \sum_{i} c_i B_i(x),
\end{equation}
where $w_b$ and $w_c $ are weights, $x/(1+\text{e}^{-x})$ is the SiLU basis function, and $\sum_{i} c_i B_i(x)$ is the linear combination of B-splines with trainable coefficients $c_i$.

The learned activation function can be set to an evaluated symbolic function with the highest R$^2$. Currently, KAN supports $x$, $x^2$, $x^3$, $x^4$, $1/x$, $1/x^2$, $1/x^3$, $1/x^4$, $\sqrt{x}$, $1/ \sqrt{x}$, $\exp(x)$, $\log(x)$, $| x |$, $\sin(x)$, $\tan(x)$, $\tanh(x)$, $\text{sigmoid}(x)$, $\text{sign}(x)$, $\arcsin(x)$, $\arctan(x)$, $\text{arctanh}(x)$, $0$, Gaussian function, $\cosh(x)$, and other self-defined functions.

\subsection{Model Performance Evaluation}

The test dataset contains data unseen by the KAN during its training. We evaluate the model's performance on this dataset in terms of the following metrics: the Nash-Sutcliffe efficiency (NSE) \cite{nash1970river}, Kling-Gupta efficiency (KGE) \cite{kling2012runoff}, RMSE, and R$^2$. Higher values of NSE, KGE, and R$^2$ signify a better model performance; lower values of RMSE signify the same. While not typically used for daily or seasonal forecasting, these statistics help us evaluate how well our models capture the overall variability in mean annual runoff components across diverse catchments.

\section{Results and Discussion}

\subsection{KAN-Derived Function $f_B$}

We use the 10-fold cross-validated grid-search method to train a KAN on the input-output pairs $\{\phi, Q_B / P\}$ extracted from the training dataset (Methods). After iterating through all hyperparameters in the grid search, the final KAN model structure is (1,1), i.e., a single learned activation function explains the relationship between $\phi$ and $Q_B / P$. In the final KAN model, the learned activation function was set as a symbolic function, and the corresponding KAN-derived $f_B$ equation is
\begin{equation} \label{eq:kan_fb}
\frac{Q_B }{ P} = f_B(\phi) = 0.39 - 0.34\, \text{tanh}(1.42\phi-0.82).
\end{equation}

\begin{figure*}[h]
\includegraphics[width=\textwidth]{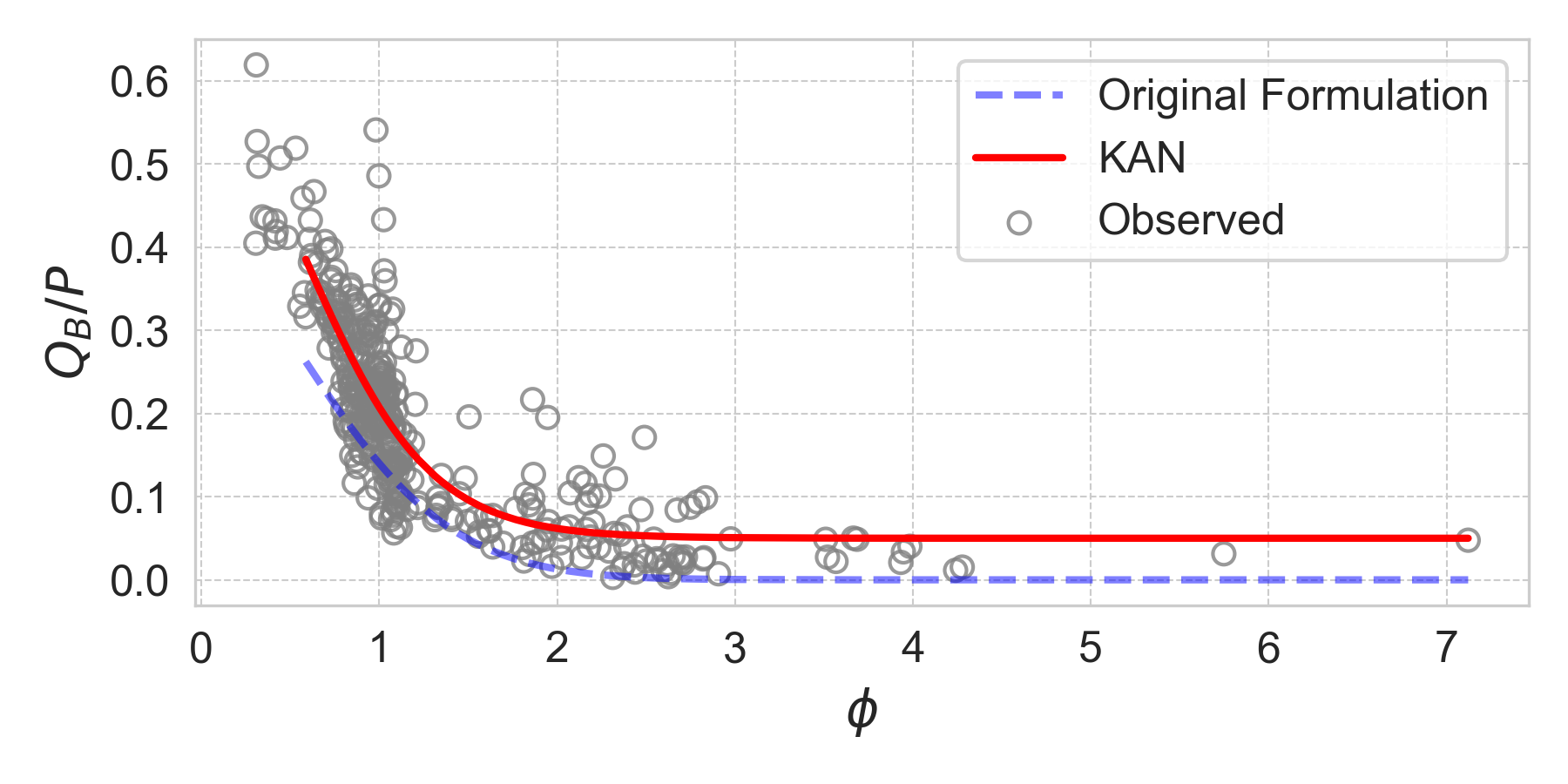}
\caption{Observed and simulated dependence of $Q_B / P$ on $\phi$ at 378 CAMELS catchments. The field observations are represented by the circles. The predictions obtained via the KAN-derived relation (Eq.~\ref{eq:kan_fb}) and the original aridity index formulation (Eq.~\ref{eq:fb_org}) are shown by the dashed and solid lines,  respectively.  }
\label{kan_eqn}
\end{figure*}

Figure \ref{kan_eqn} depicts the curves of the aridity index $f_B(\phi)$ corresponding to the KAN-derived relation (Eq.~\ref{eq:kan_fb}) and the original model (Eq.~\ref{eq:fb_org}). These are compared to the observed data $\{\phi, Q_B / P\}$ from the test dataset across 378 CAMELS catchments. The KAN-derived $f_B$ is in better agreement with the observations. Table \ref{tab:eval_metrics_kan} confirms this visual comparison. The performance of both models on the training dataset is consistent with that on the test dataset, implying that both models neither overfit nor underfit. Under both training dataset and the test dataset, all model performance metrics of the KAN-derived $f_B$ equation outperform the original aridity index formulation. The NSE increases by 71.3\%, the RMSE decreases by 32.4\%, and the KGE increases by 25\% under the test dataset.

\begin{table}[h]
    \caption{Performance metrics of the two models of $Q_B / P = f_B(\phi)$. The KAN-derived $f_B$ equation improves the model performance.}
    \label{tab:eval_metrics_kan}
    \begin{tabular*}{\textwidth}{@{\extracolsep\fill}lccccc}
        \toprule
        & \multicolumn{2}{@{}c@{}}{\textbf{Training Dataset}} & \multicolumn{3}{@{}c@{}}{\textbf{Test Dataset}} \\\cmidrule{2-3}\cmidrule{4-6}
        \textbf{Metric} & \begin{tabular}[c]{@{}c@{}}Original\\ $f_B$\end{tabular} & \begin{tabular}[c]{@{}c@{}}KAN-derived\\ $f_B$\end{tabular} & \begin{tabular}[c]{@{}c@{}}Original\\ $f_B$\end{tabular} & \begin{tabular}[c]{@{}c@{}}KAN-derived\\ $f_B$\end{tabular}  & \begin{tabular}[c]{@{}c@{}}Improvement\\ (\%)\end{tabular}  \\
        \midrule
        NSE & 0.387 & 0.747 & 0.432 & 0.741 & 71.3 \\
        KGE & 0.622 & 0.804 & 0.636 & 0.795 & 25.0 \\
        RMSE & 0.0936 & 0.0601 & 0.0866 & 0.0585 & 32.4 \\
        R$^2$ & 0.740 & 0.747 & 0.733 & 0.744 & 1.50 \\
        \hline
    \end{tabular*}
\end{table}

\subsection{KAN-Inspired $f_B$ Function }

While the KAN-derived $f_B$ equation~\eqref{eq:kan_fb} significantly outperforms the original model, it also increases the number of fitting parameters from three to four. Evaluation of the available symbolic functions used by the KAN to approximate the learned activation function revealed  $\exp(x)$, $\cosh(x)$, $\arctan(x)$, $x^4$, $\tanh(x)$ to be the top five symbolic functions, arranged in the ascending order with the corresponding R$^2$ increasing from 0.9986 to 0.9993. The exponent fits the activation function with $\text{R}^2 =  0.9986$, indicating that the original functional form of the aridity index (Eq.~\ref{eq:exp_decay}) is a reasonable choice. However, the function $\tanh(x)$ fits the activation function with $\text{R}^2 = 0.9993$, suggesting that $\tanh(x)$ might be a slightly better choice.

Although  $\exp(x)$ and $\tanh(x)$ give comparable performance in function estimation, they have different hydrological implications. The exponent is more suitable for processes that occur monotonically, such as radioactive decay. In contrast, being bounded by -1 and 1 for all $x$, $\tanh(x)$ is suitable for processes with limits, such as ET constrained by hydrological processes. This could explain why $\tanh(x)$ is a better choice for this case. It indicates underlying processes that stabilize water balance partition, exhibiting inherent limits or balancing mechanisms in the hydrological response. These considerations suggest the aridity index of the form $f_B(\phi) = a - b \tanh(\phi)$. Using the training dataset to infer fitting parameters $a$ and $b$ as
\begin{equation}  \label{eq:kan_inspired_fb}
f_B(\phi) = 0.7573 - 0.7243\tanh(\phi).
\end{equation}

Table \ref{tab:eval_metrics_new} summarizes the model performance metrics for the original aridity index formulation $f_B$ in Eq.~\eqref{eq:fb_org} and the KAN-inspired model (Eq.~\ref{eq:kan_inspired_fb}). The two models neither overfit nor underfit. The new aridity-index model shows a superior performance on both the training dataset and the test dataset,  except the 1.1\% decrease in R$^2$ for the test dataset. 
The KAN-inspired model is better than the original equation in terms of correlation, bias, and variability. 
In addition, the KAN-derived equation uses three fitting parameters, while the KAN-inspired model uses only two fitting parameters based on domain knowledge. This demonstrates that the KAN approach can effectively reduce model complexity without sacrificing performance. Both models utilize a functional form identified by KAN, highlighting KAN's ability to discover optimal functional relationships in hydrological processes. This implies that KAN can identify better functional forms to improve model accuracy, potentially leading to more physically meaningful and efficient models.

\begin{table}[h]
    \caption{Performance metrics of the two models of $Q_B / P = f_B(\phi)$. The KAN-inspired $f_B$ equation improves the model performance. }
    \begin{tabular*}{\textwidth}{@{\extracolsep\fill}lccccc}
        \toprule
        & \multicolumn{2}{@{}c@{}}{\textbf{Training Dataset}} & \multicolumn{3}{@{}c@{}}{\textbf{Test Dataset}} \\\cmidrule{2-3}\cmidrule{4-6}
        \textbf{Metric} & \begin{tabular}[c]{@{}c@{}}Original\\ $f_B$\end{tabular} & \begin{tabular}[c]{@{}c@{}}KAN-inspired\\ $f_B$\end{tabular} & \begin{tabular}[c]{@{}c@{}}Original\\ $f_B$\end{tabular} & \begin{tabular}[c]{@{}c@{}}KAN-inspired\\ $f_B$\end{tabular}  & \begin{tabular}[c]{@{}c@{}}Improvement\\ (\%)\end{tabular}  \\
        \midrule
        NSE & 0.387 & 0.745 & 0.432 & 0.720 & 66.9 \\
        KGE & 0.622 & 0.807 & 0.636 & 0.791 & 24.4 \\
        RMSE & 0.0936 & 0.0603 & 0.0866 & 0.0607 & 29.9 \\
        $\mathrm{R^2}$ & 0.740 & 0.745 & 0.733 & 0.725 & -1.10 \\
        \hline
    \end{tabular*}
    \label{tab:eval_metrics_new}
\end{table}

\subsection{Analysis of Observed Mean-Annual Variables}

\begin{figure*}
\noindent\includegraphics[width=\textwidth]{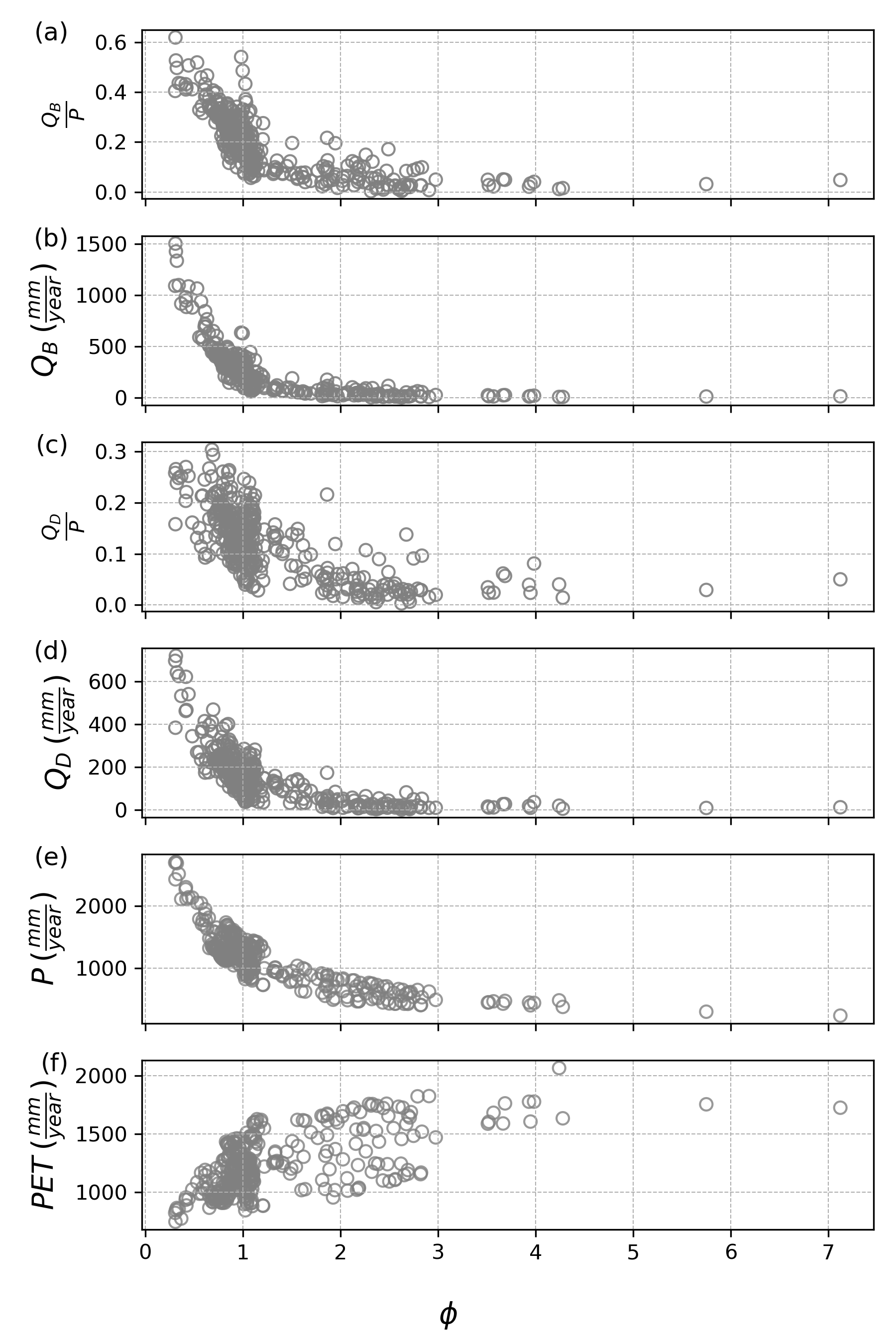}
\caption{Observed mean-annual variables plotted against the aridity index $\phi = \text{PET} / P$, for 378 CAMELS catchments.}
\label{obs_var}
\end{figure*}

Figure \ref{obs_var} exhibits the observed mean-annual variables $Q_B$, $Q_D$, $P$, and PET, along with two normalized streamflow components $Q_B / P$ and $Q_D / P$, all plotted against $\phi$ for the 378 CAMELS catchments. In accordance with \citeA{aridity}, $Q_B / P$ and $Q_D / P$ have a similar dependence on $\phi$ (Fig.~\ref{obs_var} a,c). Our analysis demonstrates that all plotted variables, with the exception of PET, decrease with $\phi$. Across the 378 CAMELS catchments, $Q_B$ and $Q_D$ show the smallest spread around the mean behavior, indicating that $\phi$ is a more reliable predictor of the variability of $Q_B$ and $Q_D$ than of their normalized counterparts $Q_B / P$ and $Q_D / P$.

We further explore the relationship between this variability and $\phi$ to uncover potential explanations. The data show that variability exists across the entire range of $\phi$, albeit with differing magnitudes. The most significant variation occurs when $\phi \in [0.5, 3]$.  This range also corresponds to significant variations in $P$ and PET (Fig.~\ref{obs_var} e-f). However, these variations are moderated by $\phi$. For instance, in catchments where a $\phi$ value approaches 1.0, $P$ varies significantly from 800 to 1500 mm/year, and PET ranges from 500 to 1500 mm/year. Such substantial variations in both $P$ and $PET$ are encapsulated by a similar $\phi$ value, contributing to a small spread of both $Q_B$ and $Q_D$ over $\phi$. $P$ and PET significantly influence the water balance as source and sink terms, respectively. Since $\phi = \text{PET} / P$, it indicates the degree to which streamflow is partitioned into direct runoff and baseflow. When these streamflow components are further normalized by $P$, additional variability from $P$ is introduced and cannot be mitigated since PET is not included in the normalization.

Based on these findings, we refine the water-balance equations~\eqref{eq:Pnorm_all} to isolate the streamflow components,
\begin{equation} \label{eq:P_all}
Q = F_D(\phi) + F_B(\phi), \qquad
F_D(\phi) \equiv Q_D, \quad 
F_B(\phi) \equiv Q_B.
\end{equation} 

The KAN-derived symbolic equations for ${F_B}$ and ${F_D}$, inferred from the training set, are
\begin{equation} \label{eq:kan_FB}
F_B(\phi) = 47.13 + 1932.52\exp(-1.42(-\phi - 0.29)^2)
\end{equation}
and
\begin{equation} \label{eq:kan_FD}
F_D(\phi) = 616.82 - 418.39\arctan(2.84\phi - 0.87).
\end{equation}
Tables~\ref{tab:fD_FD} and~\ref{tab:fB_FB} collate the performance metrics for the original aridity-index expressions for $Q_B / P$ and $Q_D / P$ and KAN-derived expressions for  $Q_B$ and $Q_D$. (We do not report RMSE because of the differing magnitudes of the target variables caused by normalization.) The KAN-derived equations significantly outperform their original counterparts in terms of all metrics. While, the KAN-derived expression for $Q_B / P$ achieves $\text{R}^2 = 0.744$ (Table \ref{tab:eval_metrics_kan}), the KAN-derived formula for $Q_B$ has $\text{R}^2 = 0.886$. This 16.4\% increase in R\textsuperscript{2} underscores the significance of refining the original water-balance equations~\eqref{eq:Pnorm_all}.

While simulating an absolute value ($Q_B$) may indeed be less challenging than simulating a ratio ($Q_B / P$), this doesn't necessarily invalidate the comparison. $Q_B$ and $Q_B / P$ have different practical applications in hydrological modeling, and comparing them provides valuable insights into model capabilities across different metrics. Our study demonstrates that KAN models for $Q_B$ can achieve significantly better prediction accuracy than models for $Q_B / P$. This finding is important as it highlights areas where our modeling approaches excel and where they may need improvement. The superior performance in $Q_B$ prediction suggests that the KAN approach is particularly effective at capturing the absolute quantity of baseflow. We acknowledge the challenges in comparing these different metrics and agree that further work is needed. Specifically, we propose to improve P prediction or incorporate additional physical constraint parameters to enhance $Q_B / P$ predictability. This approach will allow us to leverage the strengths of KAN in absolute value prediction while addressing the complexities of ratio prediction.

Finally, we compare the performance of KANs to that of two tree-based machine learning approaches: XGBoost \cite{Chen_2016} and random forest \cite{breiman2001random}. On the test dataset, the random-forest model achieves $\text{NSE} = 0.911$, $\text{KGE} = 0.941$, and $\text{R}^2 = 0.911$. The corresponding statistics for the XGBoost model is $\text{NSE} = 0.894$, $\text{KGE} = 0.854$, and $\text{R}^2 = 0.899$. The metrics of the tree-based models are similar to those of the KAN-derived relations, yet the latter are interpretable and explainable whereas the former are not. KANs generate closed-form symbolic models that are simple and can be directly used by other users without requiring any specialized software or computational tools.

\begin{table}[h]
\caption{Performance metrics of the two models of $Q_D = F_D(\phi)$. The KAN-derived $F_D$ formula improves the model performance. }
    \begin{tabular*}{\textwidth}{@{\extracolsep\fill}lccccc}
        \toprule
        & \multicolumn{2}{@{}c@{}}{\textbf{Training Dataset}} & \multicolumn{3}{@{}c@{}}{\textbf{Test Dataset}} \\\cmidrule{2-3}\cmidrule{4-6}
        \textbf{Metric} & \begin{tabular}[c]{@{}c@{}}Original\\ $f_D$\end{tabular} & \begin{tabular}[c]{@{}c@{}}KAN-derived\\ $F_D$\end{tabular} & \begin{tabular}[c]{@{}c@{}}Original\\ $f_D$\end{tabular} & \begin{tabular}[c]{@{}c@{}}KAN-derived\\ $F_D$\end{tabular}  & \begin{tabular}[c]{@{}c@{}}Improvement\\ (\%)\end{tabular}  \\
        \midrule
NSE & 0.530 & 0.769 & 0.532 & 0.768 & 44.2 \\ 
KGE & 0.604 & 0.816 & 0.601 & 0.845 & 40.6 \\ 
R\textsuperscript{2} & 0.543 & 0.769 & 0.536 & 0.769 & 43.3 \\ \hline
\end{tabular*}
\label{tab:fD_FD}
\end{table}

\begin{table}[h]
\caption{Performance metrics of the two models of $Q_B = F_B(\phi)$. The KAN-derived $F_B$ formula improves the model performance. }
    \begin{tabular*}{\textwidth}{@{\extracolsep\fill}lccccc}
        \toprule
        & \multicolumn{2}{@{}c@{}}{\textbf{Training Dataset}} & \multicolumn{3}{@{}c@{}}{\textbf{Test Dataset}} \\\cmidrule{2-3}\cmidrule{4-6}
        \textbf{Metric} & \begin{tabular}[c]{@{}c@{}}Original\\ $f_B$\end{tabular} & \begin{tabular}[c]{@{}c@{}}KAN-derived\\ $F_B$\end{tabular} & \begin{tabular}[c]{@{}c@{}}Original\\ $f_B$\end{tabular} & \begin{tabular}[c]{@{}c@{}}KAN-derived\\ $F_B$\end{tabular}  & \begin{tabular}[c]{@{}c@{}}Improvement\\ (\%)\end{tabular}  \\
        \midrule
NSE & 0.387 & 0.885 & 0.432 & 0.886 & 105 \\ 
KGE & 0.622 & 0.899 & 0.636 & 0.920 & 44.8 \\ 
R\textsuperscript{2} & 0.740 & 0.886 & 0.733 & 0.886 & 20.9 \\ \hline
\end{tabular*}
\label{tab:fB_FB}
\end{table}

\section{Conclusion}

We used KAN to derive a symbolic formula for estimating the long-term baseflow across 378 CAMELS catchments. The new equation significantly outperforms the original aridity index formula in terms of all evaluated metrics. On the test data, the NSE increases by 71.3\%, the RMSE decreases by 32.4\%, and the KGE increases by 25\%. 

KAN has identified a new ``optimal'' relation between the long-term baseflow and the aridity index. This new model with two fitting parameters outperforms the original model, which has three fitting parameters. 

Observed mean-annual variables across 378 catchments in the United States indicate that the streamflow components $Q_B$ and $Q_D$ have more robust dependencies on the aridity index $\phi$ than their normalized counterparts $Q_B / P$ and $Q_D / P$ do.

Consequently, we refine the water-balance equations to relate $Q_B$ and $Q_D$ to $\phi$. The KAN-derived symbolic relations  significantly outperform the original formulations and are comparable to the tree-based machine learning methods, highlighting the advantages of KAN in generating interpretable and accurate models for predicting streamflow components. 

Additionally, the symbolic relations identified in this study are transferable to other KAN models, allowing for the exploration of new variables or evaluation of new data by introducing more intervals to approximate complex relationships, or by investigating new activation functions while locking the activation functions identified here. These findings highlight that KAN is a promising approach to enhance hydrological modeling. 

While the aridity index can reasonably explain mean-annual streamflow components, further improvements are possible when considering other variables. For instance, hydraulic conductivity is a crucial factor in determining the rate of groundwater flow. Its spatial distribution can significantly affect the amount of baseflow. Including this parameter may further improve prediction ability. Using KAN, the learned activation function can be locked, and new activation functions related to the new input variables can be visualized and evaluated to identify potential impacts. Especially if the considered variable is not well-studied, new mechanisms between these variables may be identified. When the target function is not symbolic (e.g., Bessel function), KAN can still approximate it numerically whereas conventional symbolic regression methods may fail \cite {liu2024kan}.

\acknowledgments

This research was funded in part by the Strategic Environmental Research and Development Program (SERDP) of the Department of Defense under award RC22-3278, by the Air Force Office of Scientific Research under grant FA9550- 21-1-0381, by the National Science Foundation under award 2100927, and by the Office of Advanced Scientific Computing Research (ASCR) within the Department of Energy Office of Science under award number DE-SC0023163. The authors declare no real or perceived financial conflicts of interests.



%
%


\bibliography{kan4baseflow}

\begin{thebibliography}{}

\bibitem [\protect \citeauthoryear {%
Abatzoglou%
}{%
Abatzoglou%
}{%
{\protect \APACyear {2013}}%
}]{%
abatzoglou2013development}
\APACinsertmetastar {%
abatzoglou2013development}%
\begin{APACrefauthors}%
Abatzoglou, J\BPBI T.%
\end{APACrefauthors}%
\unskip\
\newblock
\APACrefYearMonthDay{2013}{}{}.
\newblock
{\BBOQ}\APACrefatitle {Development of gridded surface meteorological data for ecological applications and modelling} {Development of gridded surface meteorological data for ecological applications and modelling}.{\BBCQ}
\newblock
\APACjournalVolNumPages{International Journal of Climatology}{33}{1}{121--131}.
\PrintBackRefs{\CurrentBib}

\bibitem [\protect \citeauthoryear {%
Addor%
, Newman%
, Mizukami%
\BCBL {}\ \BBA {} Clark%
}{%
Addor%
\ \protect \BOthers {.}}{%
{\protect \APACyear {2017}}%
}]{%
addor2017camels}
\APACinsertmetastar {%
addor2017camels}%
\begin{APACrefauthors}%
Addor, N.%
, Newman, A\BPBI J.%
, Mizukami, N.%
\BCBL {}\ \BBA {} Clark, M\BPBI P.%
\end{APACrefauthors}%
\unskip\
\newblock
\APACrefYearMonthDay{2017}{}{}.
\newblock
{\BBOQ}\APACrefatitle {The {CAMELS} data set: catchment attributes and meteorology for large-sample studies} {The {CAMELS} data set: catchment attributes and meteorology for large-sample studies}.{\BBCQ}
\newblock
\APACjournalVolNumPages{Hydrology and Earth System Sciences}{21}{10}{5293--5313}.
\PrintBackRefs{\CurrentBib}

\bibitem [\protect \citeauthoryear {%
Allen%
, Pereira%
, Raes%
, Smith%
\BCBL {}\ \protect \BOthers {.}}{%
Allen%
\ \protect \BOthers {.}}{%
{\protect \APACyear {1998}}%
}]{%
allen1998crop}
\APACinsertmetastar {%
allen1998crop}%
\begin{APACrefauthors}%
Allen, R\BPBI G.%
, Pereira, L\BPBI S.%
, Raes, D.%
, Smith, M.%
\BCBL {}\ \BOthersPeriod {.}\end{APACrefauthors}%
\unskip\
\newblock
\APACrefYearMonthDay{1998}{}{}.
\newblock
{\BBOQ}\APACrefatitle {Crop evapotranspiration-Guidelines for computing crop water requirements --- {FAO Irrigation and Drainage Paper} 56} {Crop evapotranspiration-guidelines for computing crop water requirements --- {FAO Irrigation and Drainage Paper} 56}.{\BBCQ}
\newblock
\APACjournalVolNumPages{Fao, Rome}{300}{9}{D05109}.
\PrintBackRefs{\CurrentBib}

\bibitem [\protect \citeauthoryear {%
Assouline%
\ \BBA {} Tartakovsky%
}{%
Assouline%
\ \BBA {} Tartakovsky%
}{%
{\protect \APACyear {2001}}%
}]{%
assouline-2001-unsaturated}
\APACinsertmetastar {%
assouline-2001-unsaturated}%
\begin{APACrefauthors}%
Assouline, S.%
\BCBT {}\ \BBA {} Tartakovsky, D\BPBI M.%
\end{APACrefauthors}%
\unskip\
\newblock
\APACrefYearMonthDay{2001}{}{}.
\newblock
{\BBOQ}\APACrefatitle {Unsaturated hydraulic conductivity function based on a fragmentation process} {Unsaturated hydraulic conductivity function based on a fragmentation process}.{\BBCQ}
\newblock
\APACjournalVolNumPages{Water Resour. Res.}{37}{5}{1309-1312}.
\PrintBackRefs{\CurrentBib}

\bibitem [\protect \citeauthoryear {%
Braun%
\ \BBA {} Griebel%
}{%
Braun%
\ \BBA {} Griebel%
}{%
{\protect \APACyear {2009}}%
}]{%
Griebel}
\APACinsertmetastar {%
Griebel}%
\begin{APACrefauthors}%
Braun, J.%
\BCBT {}\ \BBA {} Griebel, M.%
\end{APACrefauthors}%
\unskip\
\newblock
\APACrefYearMonthDay{2009}{}{}.
\newblock
{\BBOQ}\APACrefatitle {On a constructive proof of {Kolmogorov}’s superposition theorem} {On a constructive proof of {Kolmogorov}’s superposition theorem}.{\BBCQ}
\newblock
\APACjournalVolNumPages{Constructive Approximation}{30}{}{653–675}.
\PrintBackRefs{\CurrentBib}

\bibitem [\protect \citeauthoryear {%
Breiman%
}{%
Breiman%
}{%
{\protect \APACyear {2001}}%
}]{%
breiman2001random}
\APACinsertmetastar {%
breiman2001random}%
\begin{APACrefauthors}%
Breiman, L.%
\end{APACrefauthors}%
\unskip\
\newblock
\APACrefYearMonthDay{2001}{}{}.
\newblock
{\BBOQ}\APACrefatitle {Random forests} {Random forests}.{\BBCQ}
\newblock
\APACjournalVolNumPages{Machine Learning}{45}{}{5--32}.
\PrintBackRefs{\CurrentBib}

\bibitem [\protect \citeauthoryear {%
Budyko%
\ \BBA {} Budyko%
}{%
Budyko%
\ \BBA {} Budyko%
}{%
{\protect \APACyear {1974}}%
}]{%
budyko1974climate}
\APACinsertmetastar {%
budyko1974climate}%
\begin{APACrefauthors}%
Budyko, M\BPBI I.%
\BCBT {}\ \BBA {} Budyko, M.%
\end{APACrefauthors}%
\unskip\
\newblock
\APACrefYear{1974}.
\newblock
\APACrefbtitle {Climate and life} {Climate and life}.
\PrintBackRefs{\CurrentBib}

\bibitem [\protect \citeauthoryear {%
Chai%
\ \BBA {} Draxler%
}{%
Chai%
\ \BBA {} Draxler%
}{%
{\protect \APACyear {2014}}%
}]{%
chai2014root}
\APACinsertmetastar {%
chai2014root}%
\begin{APACrefauthors}%
Chai, T.%
\BCBT {}\ \BBA {} Draxler, R\BPBI R.%
\end{APACrefauthors}%
\unskip\
\newblock
\APACrefYearMonthDay{2014}{}{}.
\newblock
{\BBOQ}\APACrefatitle {Root mean square error ({RMSE}) or mean absolute error ({MAE})?--{Arguments} against avoiding {RMSE} in the literature} {Root mean square error ({RMSE}) or mean absolute error ({MAE})?--{Arguments} against avoiding {RMSE} in the literature}.{\BBCQ}
\newblock
\APACjournalVolNumPages{Geoscientific Model Development}{7}{3}{1247--1250}.
\PrintBackRefs{\CurrentBib}

\bibitem [\protect \citeauthoryear {%
Chen%
\ \BBA {} Guestrin%
}{%
Chen%
\ \BBA {} Guestrin%
}{%
{\protect \APACyear {2016}}%
}]{%
Chen_2016}
\APACinsertmetastar {%
Chen_2016}%
\begin{APACrefauthors}%
Chen, T.%
\BCBT {}\ \BBA {} Guestrin, C.%
\end{APACrefauthors}%
\unskip\
\newblock
\APACrefYearMonthDay{2016}{{\APACmonth{08}}}{}.
\newblock
{\BBOQ}\APACrefatitle {{XGBoost}: A Scalable Tree Boosting System} {{XGBoost}: A scalable tree boosting system}.{\BBCQ}
\newblock
\BIn{} \APACrefbtitle {Proceedings of the 22nd {ACM SIGKDD} International Conference on Knowledge Discovery and Data Mining.} {Proceedings of the 22nd {ACM SIGKDD} international conference on knowledge discovery and data mining.}
\newblock
\APACaddressPublisher{}{ACM}.
\newblock
\begin{APACrefDOI} \doi{10.1145/2939672.2939785} \end{APACrefDOI}
\PrintBackRefs{\CurrentBib}

\bibitem [\protect \citeauthoryear {%
Head%
\ \BBA {} Zerner%
}{%
Head%
\ \BBA {} Zerner%
}{%
{\protect \APACyear {1985}}%
}]{%
head1985broyden}
\APACinsertmetastar {%
head1985broyden}%
\begin{APACrefauthors}%
Head, J\BPBI D.%
\BCBT {}\ \BBA {} Zerner, M\BPBI C.%
\end{APACrefauthors}%
\unskip\
\newblock
\APACrefYearMonthDay{1985}{}{}.
\newblock
{\BBOQ}\APACrefatitle {A {Broyden-Fletcher-Goldfarb-Shanno} optimization procedure for molecular geometries} {A {Broyden-Fletcher-Goldfarb-Shanno} optimization procedure for molecular geometries}.{\BBCQ}
\newblock
\APACjournalVolNumPages{Chemical Physics Letters}{122}{3}{264--270}.
\PrintBackRefs{\CurrentBib}

\bibitem [\protect \citeauthoryear {%
Kling%
, Fuchs%
\BCBL {}\ \BBA {} Paulin%
}{%
Kling%
\ \protect \BOthers {.}}{%
{\protect \APACyear {2012}}%
}]{%
kling2012runoff}
\APACinsertmetastar {%
kling2012runoff}%
\begin{APACrefauthors}%
Kling, H.%
, Fuchs, M.%
\BCBL {}\ \BBA {} Paulin, M.%
\end{APACrefauthors}%
\unskip\
\newblock
\APACrefYearMonthDay{2012}{}{}.
\newblock
{\BBOQ}\APACrefatitle {Runoff conditions in the upper {Danube} basin under an ensemble of climate change scenarios} {Runoff conditions in the upper {Danube} basin under an ensemble of climate change scenarios}.{\BBCQ}
\newblock
\APACjournalVolNumPages{Journal of Hydrology}{424}{}{264--277}.
\PrintBackRefs{\CurrentBib}

\bibitem [\protect \citeauthoryear {%
Kolmogorov%
}{%
Kolmogorov%
}{%
{\protect \APACyear {1956}}%
}]{%
Kolmogorov}
\APACinsertmetastar {%
Kolmogorov}%
\begin{APACrefauthors}%
Kolmogorov, A.%
\end{APACrefauthors}%
\unskip\
\newblock
\APACrefYearMonthDay{1956}{}{}.
\newblock
{\BBOQ}\APACrefatitle {On the representation of continuous functions of several variables as superpositions of continuous functions of a smaller number of variables} {On the representation of continuous functions of several variables as superpositions of continuous functions of a smaller number of variables}.{\BBCQ}
\newblock
\APACjournalVolNumPages{Dokl. Akad. Nauk}{108}{}{}.
\PrintBackRefs{\CurrentBib}

\bibitem [\protect \citeauthoryear {%
Kubalík%
, Derner%
\BCBL {}\ \BBA {} Babuška%
}{%
Kubalík%
\ \protect \BOthers {.}}{%
{\protect \APACyear {2023}}%
}]{%
Kubal_k_2023}
\APACinsertmetastar {%
Kubal_k_2023}%
\begin{APACrefauthors}%
Kubalík, J.%
, Derner, E.%
\BCBL {}\ \BBA {} Babuška, R.%
\end{APACrefauthors}%
\unskip\
\newblock
\APACrefYearMonthDay{2023}{}{}.
\newblock
{\BBOQ}\APACrefatitle {Toward Physically Plausible Data-Driven Models: A Novel Neural Network Approach to Symbolic Regression} {Toward physically plausible data-driven models: A novel neural network approach to symbolic regression}.{\BBCQ}
\newblock
\APACjournalVolNumPages{IEEE Access}{11}{}{61481–61501}.
\newblock
\begin{APACrefDOI} \doi{10.1109/access.2023.3287397} \end{APACrefDOI}
\PrintBackRefs{\CurrentBib}

\bibitem [\protect \citeauthoryear {%
Liu%
\ \protect \BOthers {.}}{%
Liu%
\ \protect \BOthers {.}}{%
{\protect \APACyear {2024}}%
}]{%
liu2024kan}
\APACinsertmetastar {%
liu2024kan}%
\begin{APACrefauthors}%
Liu, Z.%
, Wang, Y.%
, Vaidya, S.%
, Ruehle, F.%
, Halverson, J.%
, Soljačić, M.%
\BDBL {}Tegmark, M.%
\end{APACrefauthors}%
\unskip\
\newblock
\APACrefYearMonthDay{2024}{}{}.
\newblock
\APACrefbtitle {{KAN: Kolmogorov-Arnold} Networks.} {{KAN: Kolmogorov-Arnold} networks.}
\PrintBackRefs{\CurrentBib}

\bibitem [\protect \citeauthoryear {%
L'vovich%
}{%
L'vovich%
}{%
{\protect \APACyear {1979}}%
}]{%
lvovich1979world}
\APACinsertmetastar {%
lvovich1979world}%
\begin{APACrefauthors}%
L'vovich, M\BPBI I.%
\end{APACrefauthors}%
\unskip\
\newblock
\APACrefYear{1979}.
\newblock
\APACrefbtitle {World water resources and their future} {World water resources and their future}.
\newblock
\APACaddressPublisher{}{American Geophysical Union}.
\PrintBackRefs{\CurrentBib}

\bibitem [\protect \citeauthoryear {%
Lyne%
\ \BBA {} Hollick%
}{%
Lyne%
\ \BBA {} Hollick%
}{%
{\protect \APACyear {1979}}%
}]{%
lyne1979hydrology}
\APACinsertmetastar {%
lyne1979hydrology}%
\begin{APACrefauthors}%
Lyne, V.%
\BCBT {}\ \BBA {} Hollick, M.%
\end{APACrefauthors}%
\unskip\
\newblock
\APACrefYearMonthDay{1979}{}{}.
\newblock
{\BBOQ}\APACrefatitle {Hydrology and Water Resources Symposium} {Hydrology and water resources symposium}.{\BBCQ}
\newblock
\APACjournalVolNumPages{National Committee on Hydrology and Water Resources of the Institute of Engineering, Perth, Western Australia, Australia}{}{}{}.
\PrintBackRefs{\CurrentBib}

\bibitem [\protect \citeauthoryear {%
Martius%
\ \BBA {} Lampert%
}{%
Martius%
\ \BBA {} Lampert%
}{%
{\protect \APACyear {2016}}%
}]{%
martius2016extrapolation}
\APACinsertmetastar {%
martius2016extrapolation}%
\begin{APACrefauthors}%
Martius, G.%
\BCBT {}\ \BBA {} Lampert, C\BPBI H.%
\end{APACrefauthors}%
\unskip\
\newblock
\APACrefYearMonthDay{2016}{}{}.
\newblock
\APACrefbtitle {Extrapolation and learning equations.} {Extrapolation and learning equations.}
\PrintBackRefs{\CurrentBib}

\bibitem [\protect \citeauthoryear {%
Meira%
}{%
Meira%
}{%
{\protect \APACyear {2020}}%
}]{%
Meira2020}
\APACinsertmetastar {%
Meira2020}%
\begin{APACrefauthors}%
Meira, A.%
\end{APACrefauthors}%
\unskip\
\newblock
\APACrefYearMonthDay{2020}{1}{}.
\newblock
\APACrefbtitle {{Meira Neto et al 2020}.} {{Meira Neto et al 2020}.}
\newblock
\begin{APACrefDOI} \doi{10.6084/m9.figshare.11592015.v3} \end{APACrefDOI}
\PrintBackRefs{\CurrentBib}

\bibitem [\protect \citeauthoryear {%
Meira~Neto%
, Roy%
, de Oliveira%
\BCBL {}\ \BBA {} Troch%
}{%
Meira~Neto%
\ \protect \BOthers {.}}{%
{\protect \APACyear {2020}}%
}]{%
aridity}
\APACinsertmetastar {%
aridity}%
\begin{APACrefauthors}%
Meira~Neto, A\BPBI A.%
, Roy, T.%
, de Oliveira, P\BPBI T\BPBI S.%
\BCBL {}\ \BBA {} Troch, P\BPBI A.%
\end{APACrefauthors}%
\unskip\
\newblock
\APACrefYearMonthDay{2020}{}{}.
\newblock
{\BBOQ}\APACrefatitle {An Aridity Index-Based Formulation of Streamflow Components} {An aridity index-based formulation of streamflow components}.{\BBCQ}
\newblock
\APACjournalVolNumPages{Water Resources Research}{56}{9}{e2020WR027123}.
\newblock
\begin{APACrefDOI} \doi{https://doi.org/10.1029/2020WR027123} \end{APACrefDOI}
\PrintBackRefs{\CurrentBib}

\bibitem [\protect \citeauthoryear {%
Nash%
\ \BBA {} Sutcliffe%
}{%
Nash%
\ \BBA {} Sutcliffe%
}{%
{\protect \APACyear {1970}}%
}]{%
nash1970river}
\APACinsertmetastar {%
nash1970river}%
\begin{APACrefauthors}%
Nash, J\BPBI E.%
\BCBT {}\ \BBA {} Sutcliffe, J\BPBI V.%
\end{APACrefauthors}%
\unskip\
\newblock
\APACrefYearMonthDay{1970}{}{}.
\newblock
{\BBOQ}\APACrefatitle {River flow forecasting through conceptual models {part I — A} discussion of principles} {River flow forecasting through conceptual models {part I — A} discussion of principles}.{\BBCQ}
\newblock
\APACjournalVolNumPages{Journal of Hydrology}{10}{3}{282--290}.
\PrintBackRefs{\CurrentBib}

\bibitem [\protect \citeauthoryear {%
Pedregosa%
\ \protect \BOthers {.}}{%
Pedregosa%
\ \protect \BOthers {.}}{%
{\protect \APACyear {2011}}%
}]{%
pedregosa2011scikit}
\APACinsertmetastar {%
pedregosa2011scikit}%
\begin{APACrefauthors}%
Pedregosa, F.%
, Varoquaux, G.%
, Gramfort, A.%
, Michel, V.%
, Thirion, B.%
, Grisel, O.%
\BDBL {}others%
\end{APACrefauthors}%
\unskip\
\newblock
\APACrefYearMonthDay{2011}{}{}.
\newblock
{\BBOQ}\APACrefatitle {Scikit-learn: Machine learning in {Python}} {Scikit-learn: Machine learning in {Python}}.{\BBCQ}
\newblock
\APACjournalVolNumPages{The Journal of Machine Learning Research}{12}{}{2825--2830}.
\PrintBackRefs{\CurrentBib}

\bibitem [\protect \citeauthoryear {%
Ponce%
\ \BBA {} Hawkins%
}{%
Ponce%
\ \BBA {} Hawkins%
}{%
{\protect \APACyear {1996}}%
}]{%
ponce1996runoff}
\APACinsertmetastar {%
ponce1996runoff}%
\begin{APACrefauthors}%
Ponce, V\BPBI M.%
\BCBT {}\ \BBA {} Hawkins, R\BPBI H.%
\end{APACrefauthors}%
\unskip\
\newblock
\APACrefYearMonthDay{1996}{}{}.
\newblock
{\BBOQ}\APACrefatitle {Runoff curve number: Has it reached maturity?} {Runoff curve number: Has it reached maturity?}{\BBCQ}
\newblock
\APACjournalVolNumPages{Journal of Hydrologic Engineering}{1}{1}{11--19}.
\PrintBackRefs{\CurrentBib}

\bibitem [\protect \citeauthoryear {%
Sahoo%
, Lampert%
\BCBL {}\ \BBA {} Martius%
}{%
Sahoo%
\ \protect \BOthers {.}}{%
{\protect \APACyear {2018}}%
}]{%
sahoo2018learning}
\APACinsertmetastar {%
sahoo2018learning}%
\begin{APACrefauthors}%
Sahoo, S\BPBI S.%
, Lampert, C\BPBI H.%
\BCBL {}\ \BBA {} Martius, G.%
\end{APACrefauthors}%
\unskip\
\newblock
\APACrefYearMonthDay{2018}{}{}.
\newblock
\APACrefbtitle {Learning Equations for Extrapolation and Control.} {Learning equations for extrapolation and control.}
\PrintBackRefs{\CurrentBib}

\bibitem [\protect \citeauthoryear {%
Zhou%
\ \BBA {} Pan%
}{%
Zhou%
\ \BBA {} Pan%
}{%
{\protect \APACyear {2022}}%
}]{%
zhou2022bayesian}
\APACinsertmetastar {%
zhou2022bayesian}%
\begin{APACrefauthors}%
Zhou, H.%
\BCBT {}\ \BBA {} Pan, W.%
\end{APACrefauthors}%
\unskip\
\newblock
\APACrefYearMonthDay{2022}{}{}.
\newblock
\APACrefbtitle {Bayesian Learning to Discover Mathematical Operations in Governing Equations of Dynamic Systems.} {Bayesian learning to discover mathematical operations in governing equations of dynamic systems.}
\PrintBackRefs{\CurrentBib}

\end{thebibliography}

%
%
%
%
%

\end{document}